%% file: Main_for_review.tex
\documentclass[10pt,twocolumn,letterpaper]{article}

\usepackage{iccv}
\usepackage{times}
\usepackage{epsfig}
\usepackage{graphicx}
\usepackage{amsmath}
\usepackage{amssymb}
\usepackage[table]{xcolor}
\usepackage{multirow}
\usepackage{array}
\newcommand\ChangeRT[1]{\noalign{\hrule height #1}}
\usepackage{verbatim}
\usepackage{capt-of,etoolbox}
\usepackage{subcaption}

\usepackage[ruled,vlined]{algorithm2e}
\definecolor{commentcolor}{RGB}{110,154,155}   
\definecolor{controlcolor}{RGB}{219, 48, 122}

\usepackage[pagebackref=true,breaklinks=true,letterpaper=true,colorlinks,bookmarks=false]{hyperref}
\iccvfinalcopy 

\def\iccvPaperID{10847} 

\ificcvfinal\fi

\begin{document}

\title{SPANet: Frequency-balancing Token Mixer \\ using Spectral Pooling Aggregation Modulation}

\author{Guhnoo Yun$^{1,2}$ Juhan Yoo$^{3}$ Kijung Kim$^{1,2}$ Jeongho Lee$^{1,2}$ Dong Hwan Kim$^{1,2}$\\
\and 
$^1$Korea Institute of Science and Technology
\and
$^2$Korea University
$^3$Semyung University\\
{\tt\small \{doranlyong, plan100day, kape67, gregorykim\}@kist.re.kr}\\
{\tt\small unchinto@semyung.ac.kr}\\
}

\maketitle

\ificcvfinal\thispagestyle{empty}\fi

\begin{abstract}
Recent studies show that self-attentions behave like low-pass filters (as opposed to convolutions) and enhancing their high-pass filtering capability improves model performance. Contrary to this idea, we investigate existing convolution-based models with spectral analysis and observe that improving the low-pass filtering in convolution operations also leads to performance improvement. To account for this observation, we hypothesize that utilizing optimal token mixers that capture balanced representations of both high- and low-frequency components can enhance the performance of models. We verify this by decomposing visual features into the frequency domain and combining them in a balanced manner. To handle this, we replace the balancing problem with a mask filtering problem in the frequency domain. Then, we introduce a novel token-mixer named SPAM and leverage it to derive a MetaFormer model termed as SPANet. Experimental results show that the proposed method provides a way to achieve this balance, and
the balanced representations of both high- and low-frequency components can improve the performance of models on multiple computer vision tasks. Our code is available at \href{https://doranlyong.github.io/projects/spanet/}{https://doranlyong.github.io/projects/spanet/}.
\end{abstract}

\section{Introduction}
\label{sec:1_Intro}
\input{1_Intro}

\section{Related Works}
\label{sec:2_Related}

\input{2_Related}

\section{Background}
\label{sec:3_Background}
\input{3_Background}

\section{A Frequency-balancing Token Mixer}
\label{sec:4_Method}
\input{4_Method}

\section{Experiments}
\label{sec:5_Experiments}
\input{5_Experiments}

\section{Conclusion and Future Works}
\label{sec:6_Conclusion}
\input{6_Conclusion}

\section*{Acknowledgment}
\label{sec:Acknowledgment}
\input{Acknowledgment}

{\small
\bibliographystyle{ieee_fullname}
\bibliography{egbib}
}

\clearpage

\setcounter{section}{0}
\setcounter{figure}{0}
\renewcommand{\thesection}{S\arabic{section}}
\renewcommand{\thefigure}{S\arabic{figure}}

\section*{Supplementary}
\input{Supp}

\end{document}

%% file: 1_Intro.tex
\begin{figure}[t]
\vspace{0.3cm}
\centering
    \includegraphics[width=\linewidth]{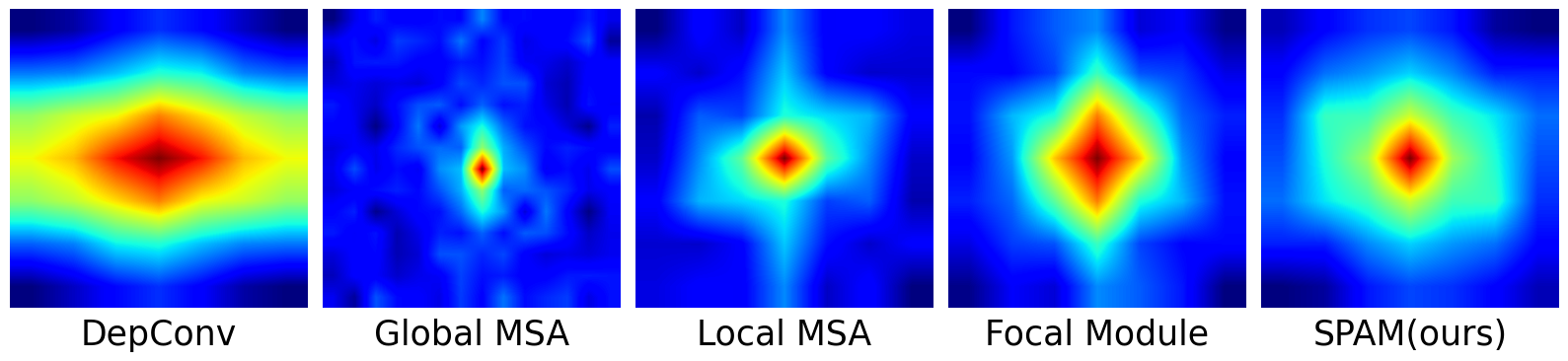}
    \caption{\textbf{Fourier spectrum maps of ConvNeXt and other MetaFormers.} The output of the spectrum map from each token mixer is processed for the same input. Depth-wise convolution (DepConv) of ConvNeXt-T~\cite{liu2022convnet}, Global MSA of ViT-B/16~\cite{dosovitskiy2020image}, Local MSA of Swin-T~\cite{liu2021Swin},  Focal module of FocalNet-T ~\cite{yang2022focal}, and SPAM of our SPANet-S are shown in order.}
    \label{fig:FFT_map}
\end{figure}

In recent years, Vision Transformers (ViTs) have achieved remarkable success and have garnered significant attention in the field of computer vision. As a result, numerous follow-up models based on the ViT~\cite{dosovitskiy2020image} have been proposed, making ViTs a dominant architecture and a viable alternative to Convolutional Neural Networks (CNNs) in various computer vision tasks including image classification~\cite{touvron2021training, wu2021cvt, liu2021Swin, vaswani2021scaling}, object detection~\cite{carion2020end, zhu2020deformable, zheng2020end}, segmentation~\cite{wang2021max, wang2021end, cheng2021per}, and beyond~\cite{chang2021augmented, zhang2021token, neimark2021video, wang2021transformer}. 

The reason for the success of ViT has been explained primarily as the use of Multi-Head Self-Attention (MSA) for token mixing~\cite{dosovitskiy2020image}. This commonly held belief has led to the development of numerous variations of MSA~\cite{d2021convit, han2021tnt, wang2021pyramid, zhou2021refiner} aimed at improving the performance of ViTs. Yet some recent works have challenged this belief by demonstrating competitive results without utilizing MSAs. Tolstikhin~\textit{et al.}~\cite{tolstikhin2021mixer} fully replaced the MSAs with a spatial Multi-Layer Perceptron (MLP) and achieves comparable results on image classification benchmarks. Subsequent studies~\cite{hou2022vision, liu2021pay, touvron2022resmlp, tang2022image} have attempted to reduce the performance gap between MLP-like models and ViTs by utilizing improved data-efficient training and redesigned MLP modules. These endeavors have shown the feasibility of MLP-like models to replace MSAs as token mixers. Moreover, other research lines~\cite{lee-thorp-etal-2022-fnet, martins2020sparse, martins2022infinite, rao2021global, han2021connection} have explored alternative self-attention-based token mixers and reported encouraging results. For example, GFNet~\cite{rao2021global} replaces self-attention with Fourier Transform and achieves competitive performance to ViT in image classification tasks. 

There have also been works that aim to understand the fundamental differences between MSAs and convolution operations. The commonly accepted explanation for the efficacy of MSAs is their capacity to effectively capture long-range dependencies without imposing a strong inductive bias~\cite{dosovitskiy2020image, naseer2021intriguing, tuli2021convolutional, yu2021rethinking, mao2022towards, chu2021Twins} in contrast to convolution operations. In a recent study, however, Park~\textit{et al.}~\cite{park2022how} explored the spectral filtering properties of both MSAs and convolutions and found that MSAs are closer to low-pass filtering, while convolution operations are better suited for filtering high-pass signals. The study also suggests that incorporating both operations in a specific sequence can lead to improved performance. Another study done by Bai~\textit{et al.}~\cite{bai2022improving} investigates the adversarial robustness of MSAs and convolution operations by adding frequency perturbation and reached a similar conclusion. Moreover, the study proposes three training schemes to enhance the capture of high-frequency components by MSAs, leading to performance improvement of ViTs. That is, the model performance can be improved by enhancing the weak high-pass filtering capability of MSAs, or by using a token mixer optimized from a spectral filter perspective. Conversely, it can be expected that enhancing the low-pass filter capability of convolutions can also improve performance.

\begin{figure}[t]
\vspace{0.3cm}
\centering
    \includegraphics[width=0.78 \linewidth]{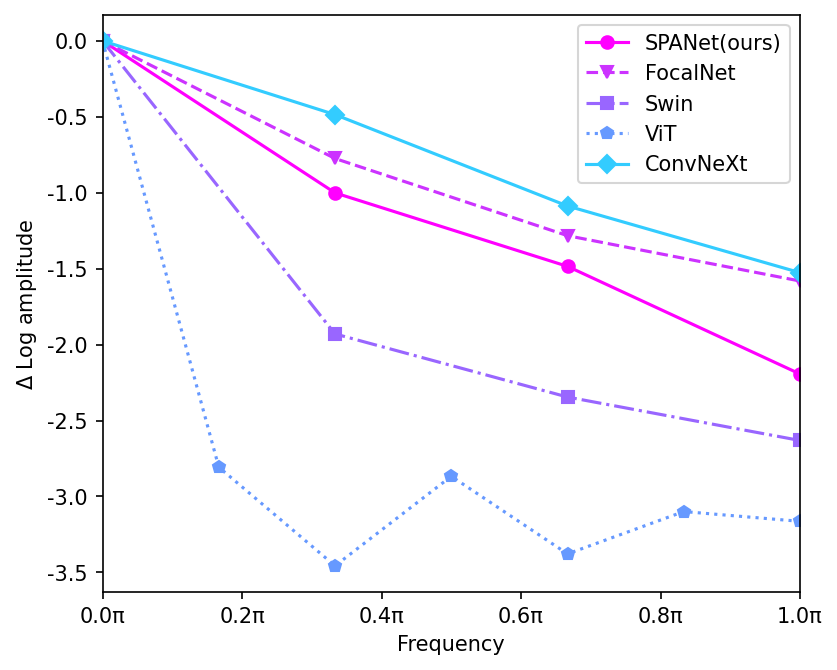}
    \caption{\textbf{Relative log amplitude of Fourier transformed feature maps.} DepConv of ConvNeXt-T and Global MSA of ViT-B/16  exhibit low-pass filtering that contains the most and the least high-frequency components, respectively. Conversely, Local MSA of Swin-T and Focal Module of FocalNet-T seem to capture both spectral components in a better balanced way.}
    \label{fig:log_amp}
\end{figure}

Figures~\ref{fig:FFT_map} and~\ref{fig:log_amp} provide evidence supporting the expectation. Consistent with previous studies, depth-wise convolution (DepConv) is relatively more effective at capturing high-frequency signals compared to local- and global-MSA. On the other hand, the Focal Module~\cite{yang2022focal} demonstrates better low-pass filtering capability, despite utilizing DepConv, and its performance also surpasses that of ConvNeXt~\cite{liu2022convnet}, ViT~\cite{dosovitskiy2020image}, and Swin Transformer~\cite{liu2021Swin}. Collecting all these results together, we then naturally make such a hypothesis: \textit{utilizing optimal token mixers that capture balanced representations of both high- and low-frequency components can enhance the performance of models}.

To verify this hypothesis, we employ the Discrete Fourier Transform (DFT) to decompose visual features into low- and high-frequency components. We then assign weights to tokens corresponding to each frequency band to balance low-frequency and high-frequency components in a way. To accomplish this, we replace the balancing problem with a mask filtering problem in the frequency domain and introduce a novel token-mixer called \textit{spectral pooling aggregation modulation} (SPAM) module, which enables the balance of high- and low-frequency components. Using the SPAM token-mixer, we propose \textit{SPANet} based on the MetaFormer architecture~\cite{yu2022metaformer}. The performance of SPANet is evaluated on three benchmark computer vision tasks: image classification, object detection, and segmentation, and it demonstrates improved results compared to the previous state-of-the-art. 

Our contributions are summarized as three-fold. (1) We handle the balancing problem of high- and low-frequency components of visual features, and show that it can be replaced with a mask filtering problem in the frequency domain. Specifically, we solve this problem by introducing SPAM. (2)  Leveraging SPAM, we propose SPANet, which is based on the MetaFormer architecture~\cite{yu2022metaformer}. (3)  Our proposed SPANet is evaluated on multiple vision tasks, including image classification~\cite{deng2009imagenet}, object detection~\cite{lin2014microsoft}, instance segmentation~\cite{lin2014microsoft}, and semantic segmentation~\cite{zhou2017scene}. Our results show that SPANet outperforms state-of-the-art models.

%% file: 2_Related.tex
\subsection{Transformers}
Transformer has been first proposed in ~\cite{vaswani2017attention} for machine language translation which utilizes self-attention to learn representations of the input sequence that capture long-range dependencies and relationships between different language tokens. Thanks to its successful application in many natural language processing (NLP) tasks, the applicability of self-attention has been extended to the computer vision field. For instance, ViT~\cite{dosovitskiy2020image} pioneered how to adopt a pure transformer architecture in image classification tasks and achieve excellent performance. Since the success of ViT, many follow-up works have been focusing on improving the MSA-based token mixers of ViTs through various approaches, such as shifted windows~\cite{liu2021Swin}, relative position encoding~\cite{wu2021rethinking}, anti-aliasing attention map~\cite{qian2021blending}, or incorporating convolution~\cite{d2021convit, guo2022cmt, wu2021cvt}, \textit{etc}.

\subsection{MetaFormers beyond Self-Attentions}
Despite the widespread belief that the MSAs play an essential role in the success of ViTs, some recent studies have raised the question of whether it is the crucial element responsible for their high performance. For instance, it was found that MSAs can be entirely substituted with MLPs as token mixers~\cite{tolstikhin2021mixer, touvron2022resmlp}, while still achieving competitive performance relative to ViTs. This discovery sparked a discussion in the research community about which token mixer is better~\cite{chen2022cyclemlp, hou2022vision} and several works challenged the dominance of attention-based token mixers by replacing MSAs with various approaches~\cite{lee-thorp-etal-2022-fnet, martins2020sparse, martins2022infinite, rao2021global}. Meanwhile, there have been other studies to explore transformers from the aspect of general architecture termed MetaFormer by replacing MSAs with non-parametric token mixers. ShiftViT~\cite{wang2022shift} uses a partial shift operation~\cite{lin2019tsm} instead of MSAs, and PoolFormer~\cite{yu2022metaformer} employs a spatial average pooling operator to replace MSAs. Both models achieve competitive performance on various computer vision tasks, suggesting that utilizing MetaFormer architecture can lead to reasonable performance. Building on this idea, we propose SPANet leveraging the advantage of MetaFormer architecture.


\subsection{Frequency Domain Analysis}
The frequency domain analysis has been extensively studied in the literature on computer vision. Normally, the low frequencies correspond to global structures and color information while the high frequencies correspond to fine details of objects (~\textit{e.g.}, local edges/textures)~\cite{cooley1969fast, deng1993adaptive}. According to ~\cite{park2022how, bai2022improving}, MSAs highly tend to learn low-frequency representations in visual data but are weak for learning high-frequencies. On the other hand, convolutions exhibit the opposite behavior. Based on these observations, LITv2~\cite{pan2022hilo} proposed a HiLo attention-mixer which captures both high- and low-frequency information with self-attention. Furthermore, Bai~\textit{et al.}~\cite{bai2022improving} proposed HAT that enhances the ability of ViTs to capture high-frequency components using adversarial training. To the best of our knowledge, however, there has been no prior work aimed at enhancing CNNs in effectively capturing low-frequency components in visual data. Inspired by this, we introduce a new token-mixer called SPAM, which utilizes convolutional operation to efficiently capture both high- and low-frequency signals in a balanced manner.

%% file: 3_Background.tex
\subsection{Feature Filtering in the Frequency Domain}
Typically, there are two types of methods for image filtering. One is to perform a kernel convolution in the spatial domain and the other is to utilize the Discrete Fourier Transform (DFT) for filtering in the frequency domain. According to the convolution theorem~\cite{katznelson2004introduction}, the results of visual feature processing in either the spatial domain or the frequency domain are equivalent. Yet transforming the features into the frequency domain allows for direct control of the spectral signals of features. Therefore, we adopt the frequency-based filtering method using the 2D DFT. This process is divided into three steps as follows.

Given a visual feature $\boldsymbol{x} \in \mathbb{R}^{H \times W \times D}$ as input, 2D DFT is used to transform it from the spatial domain to the frequency domain:
\begin{equation} \label{eq_xc}
\boldsymbol{X}_{c} = \mathcal{F}(\boldsymbol{x}_{c}) \in \mathbb{C}^{H \times W},
\end{equation} 
where $\mathcal{F}(\cdot)$ denotes 2D DFT function, $\boldsymbol{x}_{c} \in \mathbb{R}^{H \times W}$ represents the $c$-th dimension of visual feature $\boldsymbol{x}$, and $\boldsymbol{X}_{c}$ is a complex tensor representing the spectrum of $\boldsymbol{x}_{c}$. We use \texttt{torch.fft.fft2} implemented by PyTorch library~\cite{paszke2019pytorch} to apply $\mathcal{F}(\cdot)$ to $\boldsymbol{x}_{c}$.    

The desired frequency band is then modified by applying the Hadamard product (HP) with a weighting matrix to weight the spectrum:
\begin{equation} \label{eq_xct}
\boldsymbol{\tilde{X}}_{c} = \boldsymbol{M} \odot \boldsymbol{X}_{c},
\end{equation} 
where $\odot$ denotes the HP and $\boldsymbol{M}$ is an arbitrary weighting matrix that has the same size as $\boldsymbol{X}_{c}$.

Finally, the inverse DFT is applied to convert the modulated $\boldsymbol{\tilde{X}}_{c}$ back into the spatial domain and update the features:
\begin{equation} \label{eq_bsd}
\boldsymbol{x}_{c}\leftarrow \boldsymbol{\tilde{x}}_{c}= \mathcal{F}^{-1}(\boldsymbol{\tilde{X}}_{c}).
\end{equation}

\subsection{Focal Modulation}
The focal modulation~\cite{yang2022focal} is a new method that exploits depth-wise convolution to mimic the self-attention in a different way. This approach first aggregates context features, then interacts with visual tokens using the HP as: 
\begin{equation} \label{eq_ha}
\boldsymbol{y}^{k} = q(\boldsymbol{x}^{k}) \odot m(k,\boldsymbol{x}),
\end{equation} 
where $\boldsymbol{x}^{k} \in \mathbb{R}^{D}$ is visual token (query) at position $k$ and $\boldsymbol{y}^{k} \in \mathbb{R}^{D}$ is refined representation. $q(\cdot)$ and $m(\cdot)$ are functions for query projection and context aggregation, respectively. 

By observing Figures~\ref{fig:FFT_map} and~\ref{fig:log_amp}, the transformed feature of the focal modulation has a relatively more concentration of low-frequency signals compared to that of the DepConv. This result suggests that modulation with $m(\cdot)$ has a structural advantage for constructing a low-pass filter. Motivated by this, we leverage the focal modulation strategy described in Eq.~\ref{eq_ha} for our token-mixer design.


%% file: 4_Method.tex
In this section, we introduce a novel context aggregation using convolutional modulation. Since convolution operations tend to relatively favor high-pass filtering~\cite{park2022how}, we aim to modulate the context features to concentrate relatively more on the low-pass signal for balance.


\subsection{Spectral Pooling Gate (SPG)}
For simplicity of implementation, we decompose a visual feature into a combination of low-pass ($\textit{lp}$) and high-pass ($\textit{hp}$) filters. That is, the low- and high-frequency components from the input visual features $\boldsymbol{x}$  are filtered out by pre-defined filters and then blended into one. This can be expressed in the following equation:
\begin{equation} \label{eq_pdf}
\boldsymbol{\tilde{x}}_{c} =\lambda_{b} f_{lp}(\boldsymbol{x}_{c})+(1-\lambda_{b})f_{hp}(\boldsymbol{x}_{c}) \in \mathbb{R}^{H \times W},
\end{equation} 
where $\lambda_{b} \in [0,1]$ is a balancing parameter and $\boldsymbol{\tilde{x}}_{c}$ represents the filtered $\boldsymbol{x}_{c}$ by the combination of low- and high-pass filters. 

Now the balance of the high- and low-frequency components can be controlled by manipulating the spectrum of the visual features by adjusting $\lambda_{b}$. For example, setting $\lambda_{b}$ to 0.5,  the output after normalization will be the same as the normalized input without any transformation.


\subsubsection{Filtering with Spectral Pooling Filter (SPF)}
Spectral pooling introduced by Rippel~\textit{et al.}~\cite{rippel2015spectral} is a pooling technique that is used to reduce spatial tensor dimension by applying a low-pass filter. 
This is based on the inverse power law, which states that the expected power of natural images is statistically concentrated in the low-frequency region~\cite{torralba2003statistics}. In other words, most of the important visual information in natural images is contained in the low-frequency part of the spectrum. Based on this, we design that low-frequency components are given greater weight compared to high-frequency components for frequency balancing. Also, it is general to preserve the input and output dimensions in traditional token-mixer designs. In the proposed spectral pooling scheme, therefore, filtering is applied while preserving the dimension.

The first step is to apply the 2D DFT to the input feature map and shift it so that the low-frequency components are located at the center(\textit{i.e.}, the origin is set in the middle of the spectral map). 
For the low pass filter, $f_{lp}$, we select a low-frequency subset and remove the rest as follows: 
\begin{equation} \label{eq_slfcalf}
\boldsymbol{S}^{lf}_{c}=\begin{cases}\mathcal{G}(\boldsymbol{X}_{c})(u,v) & (u,v) \in \mathbf{A}^{lf} \\ 0 & \text{otherwise}\end{cases},
\end{equation} 
where $\mathcal{G}(\cdot)$ is a function for centering the Fourier transform (we use \texttt{torch.fft.fftshift} implemented by PyTorch~\cite{paszke2019pytorch} library), $(u,v)$ is a pair of positions for frequency-domain, and $\mathbf{A}^{lf} \in \mathbb{R}^{2}$ is a selected low-frequency region centered on the origin.
Then, we obtain the spectral pooled feature map by applying the inverted shift and the inverse DFT:
\begin{equation} \label{eq_flp}
f_{lp}(\boldsymbol{x}_{c}) =\mathcal{F}^{-1}(\mathcal{G}^{-1}(\boldsymbol{S}^{lf}_{c}))\in \mathbb{R}^{H \times W}.
\end{equation} 

The high-pass filter, $f_{hp}$, acts in the opposite manner to the low-pass filter and can be obtained by blocking or subtracting low-frequency components from the input feature map as follows:
\begin{equation} \label{eq_lf}
\boldsymbol{S}^{hf}_{c} = \mathcal{G}(\boldsymbol{X}_{c}) - \boldsymbol{S}^{lf}_{c},
\end{equation} 
where $\boldsymbol{S}^{hf}_{c} \in \mathbb{C}^{H \times W}$ is the high-frequency subset with the low-frequency area $\mathbf{A}^{lf}$ filled with zeros in $\mathcal{G}(\boldsymbol{X}_{c})$.
Subsequently, the inverse DFT is applied to the inverted shift of the high-frequency subset in a similar fashion as in Eq.~\ref{eq_flp} to obtain the high-pass filtered outcome:
\begin{equation} \label{eq_fhp}
f_{hp}(\boldsymbol{x}_{c}) =\mathcal{F}^{-1}(\mathcal{G}^{-1}(\boldsymbol{S}^{hf}_{c}))\in \mathbb{R}^{H \times W}.
\end{equation}

\subsubsection{Implementation of SPF using Mask Filtering}
\label{sec_Implementation}
Since $\mathcal{F}$, $\mathcal{G}$, and those inverses are linear systems, they satisfy the superposition property. Therefore, Eq.~\ref{eq_pdf} can be replaced by using Eq.~\ref{eq_flp} and Eq.~\ref{eq_fhp} as follows: 
\begin{equation} \label{eq_txc}
\boldsymbol{\tilde{x}}_{c}=\mathcal{F}^{-1}(\mathcal{G}^{-1}(\lambda_{b}\boldsymbol{S}^{lf}_{c}+(1-\lambda_{b})\boldsymbol{S}^{hf}_{c})).
\end{equation} 

In fact, the process to obtain spectral-pooled subsets ($\boldsymbol{S}^{lf}_{c}$ and $\boldsymbol{S}^{hf}_{c}$) by cropping the target band and filling the rest with zeros, can be easily achieved by masking the spectral map $\mathcal{G}(\boldsymbol{X}_{c})$ with ideal binary masks using Eq.~\ref{eq_xct}. The binary mask $\boldsymbol{M}^{lf}$ for obtaining $\boldsymbol{S}^{lf}_{c}$ is filled with ones in $\mathbf{A}^{lf}$ and zeros in the rest as follows:
\begin{equation} \label{eq_mlf}
\boldsymbol{M}^{lf}=\begin{cases} 1& (u,v) \in \mathbf{A}^{lf} \\ 0& \text{ otherwise } \end{cases}.
\end{equation} 
Conversely, the binary mask $\boldsymbol{M}^{hf}$ is obtained with filling zeros in $\mathbf{A}^{lf}$ and ones in the rest:
\begin{equation} \label{eq_mhf}
\boldsymbol{M}^{hf}=\begin{cases} 0& (u,v) \in \mathbf{A}^{lf} \\ 1& \text{ otherwise } \end{cases}.
\end{equation} 
Now the spectral-pooled subsets, $\boldsymbol{S}^{lf}_{c}$ and $\boldsymbol{S}^{hf}_{c}$, can be obtained by simple mask operation as follows:
\begin{equation} \label{eq_slfc}
\boldsymbol{S}^{lf}_{c}=\boldsymbol{M}^{lf} \odot \mathcal{G}(\boldsymbol{X}_{c}),
\end{equation} 
\begin{equation} \label{eq_shfc}
\boldsymbol{S}^{hf}_{c}=\boldsymbol{M}^{hf} \odot \mathcal{G}(\boldsymbol{X}_{c}).
\end{equation} 
Thus, $\lambda_{b}\boldsymbol{S}^{lf}_{c}+(1-\lambda_{b})\boldsymbol{S}^{hf}_{c}$ can be described as below by applying Eq.~\ref{eq_slfc} and Eq.~\ref{eq_shfc}: 
\begin{equation} \label{eq_ee}
(\lambda_{b}\boldsymbol{M}^{lf}+(1-\lambda_{b})\boldsymbol{M}^{hf}) \odot \mathcal{G}(\boldsymbol{X}_{c}).
\end{equation}

Any filter can be described by combining two or more ideal filters. In Eq.~\ref{eq_ee}, $\lambda_{b}\boldsymbol{M}^{lf}$ means scaling the values in $\mathbf{A}^{lf}$ by $\lambda_{b}$, and $(1-\lambda_{b})\boldsymbol{M}^{hf}$ means scaling the values except $\mathbf{A}^{lf}$ by $(1-\lambda_{b})$. For efficient mask operation, therefore, $\lambda_{b}\boldsymbol{M}^{lf}+(1-\lambda_{b})\boldsymbol{M}^{hf}$ can be combined as a single mask:
\begin{equation} \label{eq_mfa}
\boldsymbol{M}^{f}=\begin{cases} \lambda_{b}& (u,v) \in \mathbf{A}^{lf} \\ 1-\lambda_{b}& \text{ otherwise } \end{cases},
\end{equation} 
where $\boldsymbol{M}^{f} \in \mathbb{R}^{H\times W}$ is the combination of $\boldsymbol{M}^{lf}$ and $\boldsymbol{M}^{hf}$. 
Therefore, Eq.~\ref{eq_txc} is simply rewritten as:
\begin{equation} \label{re_eq_txc}
\boldsymbol{\tilde{x}}_{c}=\mathcal{F}^{-1}(\mathcal{G}^{-1}(\boldsymbol{M}^{f}\odot \mathcal{G}(\boldsymbol{X}_{c})).
\end{equation}

Finally, we need to define $\mathbf{A}^{lf}$ of Eq.~\ref{eq_slfcalf} in detail. In the spectral pooling of Rippel~\textit{et al.}~\cite{rippel2015spectral}, $\mathbf{A}^{lf}$ is described as a rectangular shape. Generally, rectangular low-pass filtering, however, can result in artifacts or distortion in the output image. Therefore, we define $\mathbf{A}^{lf}$ as a circular shape:
\begin{equation} \label{eq_alfuv}
\mathbf{A}^{lf}(u,v)=\{(u,v)|\sqrt{(u-u_0)^2+(v-v_0)^2}<r\},
\end{equation} 
where $(u_{0}, v_{0})$ indicates the origin of $(u,v)$ pairs and $r$ is a radius. That is, $\lambda_{b}$ is assigned to the locations within radius $r$ and $1-\lambda_{b}$ is assigned to the rest.

\begin{figure}[!tp]
\vspace{0.3cm}
\centering
    \includegraphics[width=0.9\linewidth]{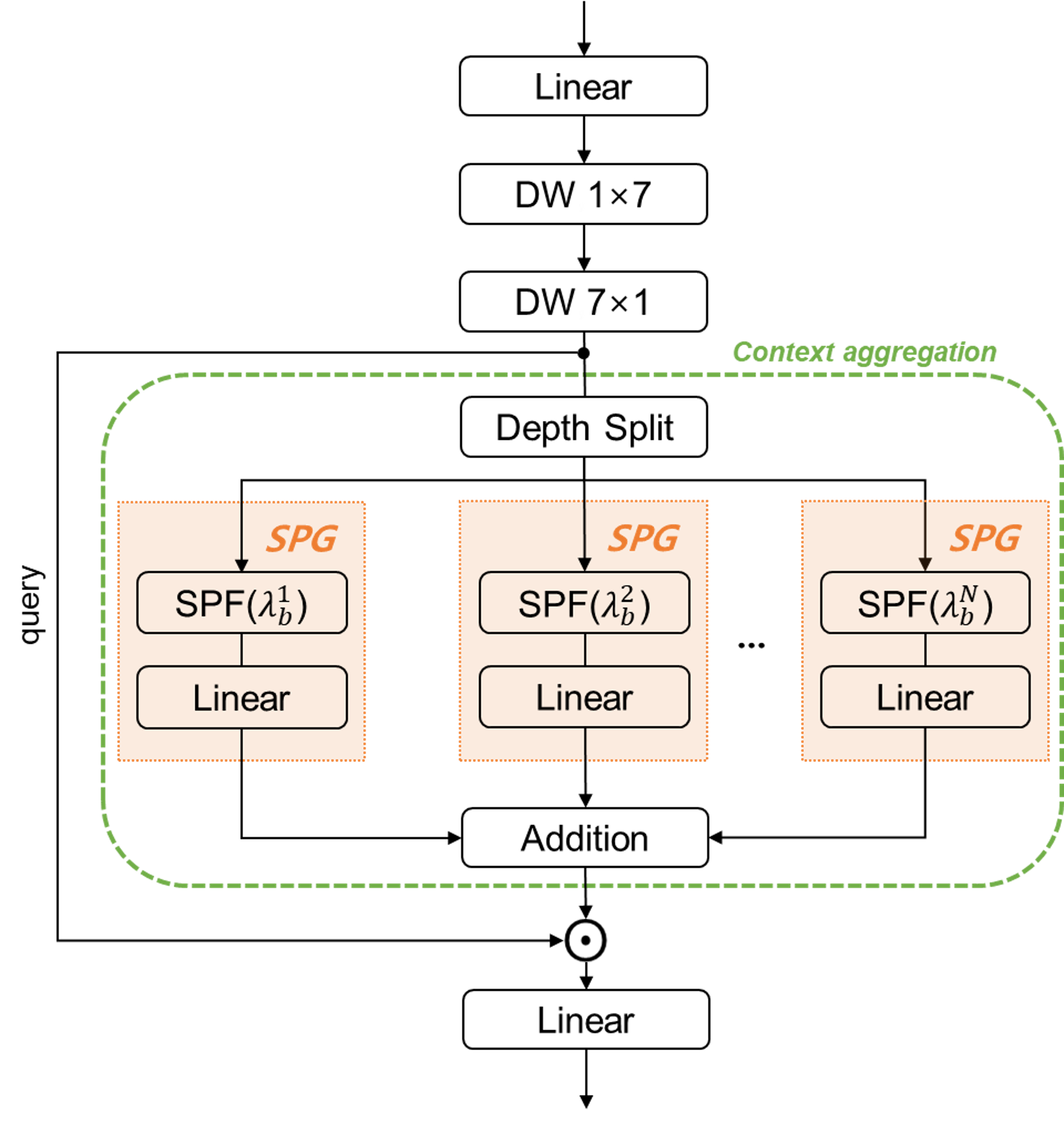}
    \caption{\textbf{Overview of the SPAM.} DW represents a depth-wise convolution and the block 'Linear' is implemented using a $1\times1$ convolution.}
    \label{fig:SPAM}
\end{figure}

\subsubsection{Feature Interaction}
Applying the pre-defined filter in Section~\ref{sec_Implementation} uniformly to all feature dimensions is generally simplistic but limits the ability to reliably optimize representations by considering correlations between feature maps. In order to deal with this problem, Qian~\textit{et al.}~\cite{qian2021blending} derived various and complex filters from the pre-defined filters with a linear assembling strategy with $1\times1$ convolutions. In this paper, we also apply the same scheme using Eq.~\ref{re_eq_txc} :
\begin{equation} \label{eq_xi}
\boldsymbol{x}_{i} = \sum_{c=1}^{D}\phi_{i,c}\boldsymbol{\tilde{x}}_{c},
\end{equation}
where  $\phi_{i,c} \in \mathbb{R}$ denotes the $c$-th learnable parameter of $i$-th kernel of $1\times1$ convolutions, and $\boldsymbol{x}_{i} \in \mathbb{R}^{H \times W}$ is a dynamically interacted feature map of $\boldsymbol{\tilde{x}} \in \mathbb{R}^{H \times W \times D}$.

As a result, SPG adjusts the high- and low-frequency components of all visual features using SPF of Eq.~\ref{re_eq_txc} and expresses complex and rich features utilizing Eq.~\ref{eq_xi}, while optimizing the balance of frequency components. The overview of SPG is included in Figure~\ref{fig:SPAM}.

\subsection{Spectral Pooling Aggregation Modulation}
\begin{figure}[!tp]
\vspace{0.3cm}
\centering
    \includegraphics[width=0.9\linewidth]{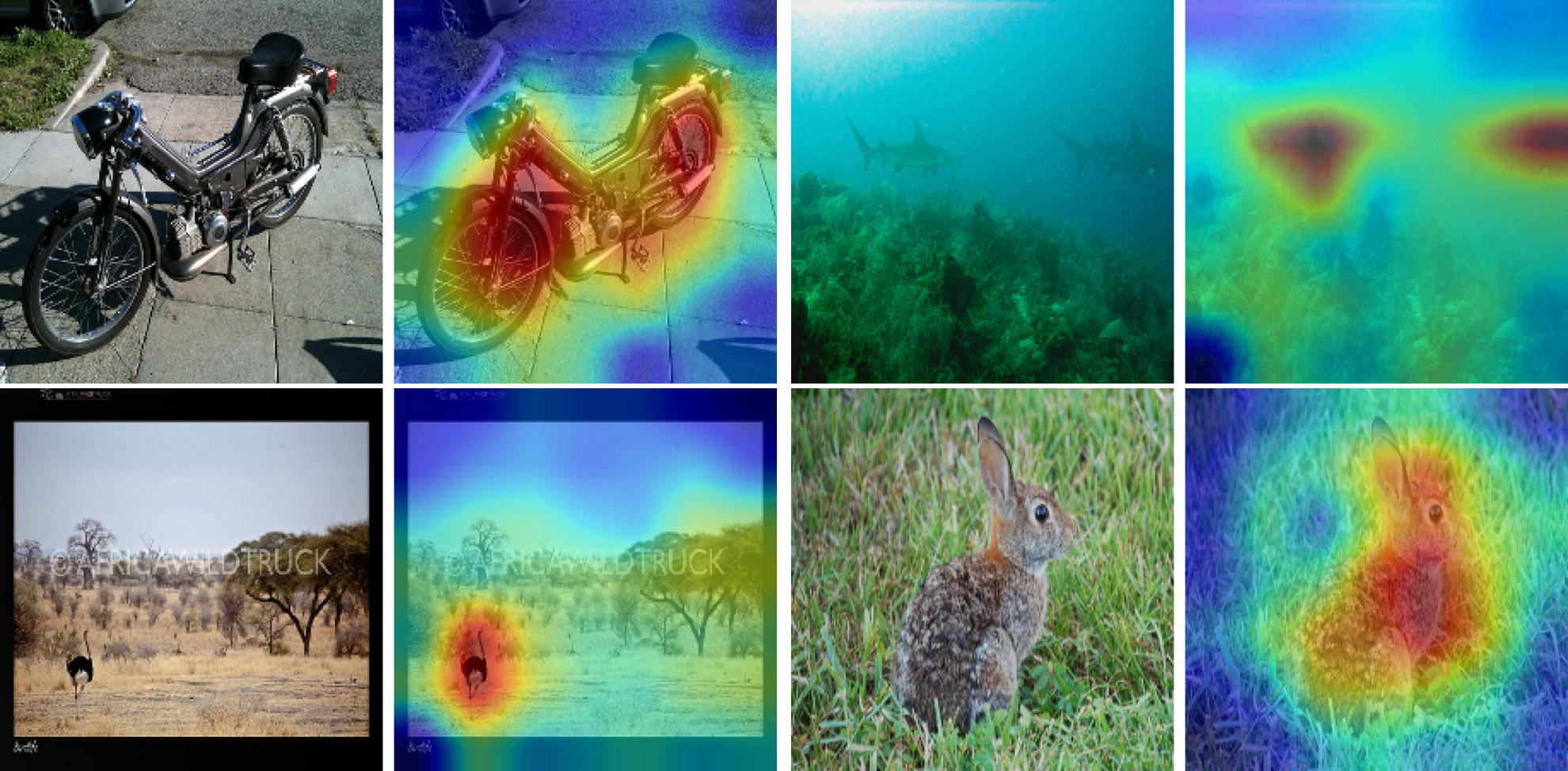}
    \caption{\textbf{Visualization of the context map.} The context map derived from the aggregated SPG features appropriately aligns with the object in the given image. This demonstrates that the aggregated context of SPAM can exhibit interpretable contextual features without self-attentions.}
    \label{fig:ctx_map}
\end{figure}

In this section, we propose a novel context aggregation using SPG. We then introduce a new token-mixer called \textit{Spectral Pooling Aggregation Modulation} (SPAM) following the same strategy in Eq.~\ref{eq_ha}. The overall structure is shown in Figure~\ref{fig:SPAM}. Given a visual feature $\boldsymbol{x}$, it passes through a linear layer and depth-wise convolution for query projection. To reduce the number of parameters, spatial separable convolution~\cite{szegedy2016rethinking} is adopted, which decompose $K \times K$ kernel into a pair of $1\times K$ and $K\times1$. In the context aggregation phase, $N$ SPGs are utilized to aggregate filtered values by various balancing parameters. Each SPG receives a uniformly split projection map, and its output is aggregated by addition for context. The context map is shown in Figure~\ref{fig:ctx_map}. Then, the aggregated context is applied to the query for modulation. Finally, the modulated feature is passed through a linear layer for interaction.

\begin{table}[t]
\centering
\caption{\textbf{Model configurations of SPANets.} $C$, $L$, and $r$ mean embedding dimension, layer number (as known as depth), and radius of SPF in each stage, respectively. Each row describes each model variant for small, medium, and base denoted as S, M, and B, respectively.}
\begin{tabular}{l|c|c|c|c}
\ChangeRT{0.8pt}
\hline
Model                   & size & $C$              & $L$        & $r$       \\ \hline
\multirow{3}{*}{SPANet} & S    & 64-128-320-512 & 4-4-12-4 & 2-2-1-1   \\ \cline{2-5} 
                        & M    & 64-128-320-512 & 6-6-18-6 & 2-2-1-1   \\ \cline{2-5} 
                        & B    & 96-192-384-768 & 6-6-18-6 & 2-2-1-1  \\ \ChangeRT{0.8pt} \hline
\end{tabular}%
\label{table:ablation_T}
\end{table}

\subsection{SPANet Architectures}
We adopt the same stage layouts and embedding dimensions as in the MetaFormer baseline~\cite{yu2022metaformer} but replace the token-mixer parts with the proposed SPAM to construct a series of \textit{SPAM Network} (SPANet) variants. In SPANets, we only need to specify the balancing parameters, $\lambda_{b}$, for each SPG, along with the radius, $r$, for the low-frequency band at each stage. The detailed configurations for each variant labeled as small, medium, and base are described in Table~\ref{table:ablation_T}. Following the inverse power law~\cite{torralba2003statistics}, we assume $\lambda_{b}$ should be larger than 0.5 to  emphasize low-frequency components. Experimentally, we set $N$ to 3, and $\lambda_{b}$ of each SPG to 0.7, 0.8, and 0.9, respectively.

%% file: 5_Experiments.tex
Following common practices~\cite{yu2022metaformer, liu2021Swin, yang2021focal, wang2021pyramid}, we conduct experiments to verify the effectiveness of the proposed SPANet on three tasks: image classification on ImageNet-1K~\cite{deng2009imagenet}, object detection and instance segmentation on COCO~\cite{lin2014microsoft} and semantic segmentation on ADE20K~\cite{zhou2017scene}. Firstly, we evaluate the proposed SPANet architecture against the previous state-of-the-art on three tasks. In addition, the ablation study section analyzes the significance of the design elements of the proposed architecture. All experiments were implemented using PyTorch~\cite{paszke2019pytorch} on Ubuntu 20.04 with 4 NVIDIA RTX3090 GPUs.

\begin{figure}[!tp]
\vspace{0.3cm}
\centering
    \includegraphics[width=0.89 \linewidth]{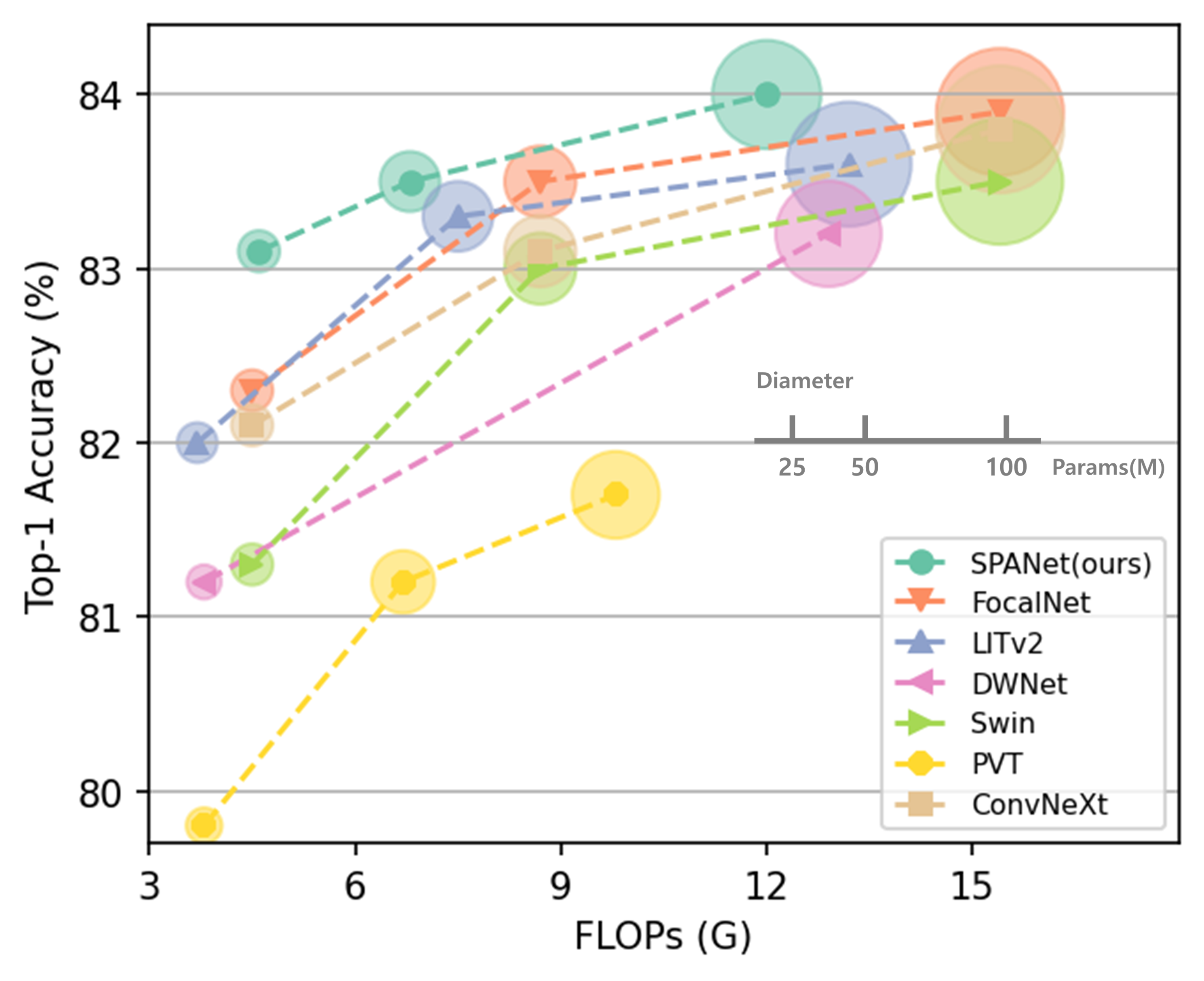}
    \caption{\textbf{ImageNet-1K validation accuracy vs. FLOPs/Params for SPANets and other comparative models.} The size of each bubble is proportional to the number of parameters in a variant within a model family.}
    \label{fig:acc_flops}
\end{figure}

\subsection{Image Classification on ImageNet-1K}

\textbf{Implementation setup.} For image classification, we evaluated SPANet on ImageNet-1K~\cite{deng2009imagenet} which is one of the most widely cited benchmarks in the computer vision society. It comprises 1.28M training images and 50K validation images from 1K classes. Most of the training strategies are followed in ~\cite{yu2022metaformer} and ~\cite{touvron2021training}. The models are trained for 300 epochs at $224^{2}$ resolution by AdamW optimizer~\cite{kingma2014adam, loshchilov2017decoupled} with weight decay $0.05$ and peak learning rate lr$=1e^{-3} \times \frac{\textrm{batch size}}{1024}$ (a batch size of 1024 and a learning rate of $1e^{-3}$ are used in this paper). The number of warmup epochs is 5 and a cosine decay learning rate scheduler is used. For data augmentation and regularization, MixUp~\cite{zhang2018mixup}, CutMix~\cite{yun2019cutmix}, CutOut~\cite{zhong2020random}, RandAugment~\cite{cubuk2020randaugment}, Label Smoothing~\cite{szegedy2016rethinking} and Stochastic Depth~\cite{huang2016deep} are used. Dropout is disabled but ResScale~\cite{shleifer2021normformer} for the last two stages is adopted to aid in training deep models. We employed Modified Layer Normalization (MLN)~\cite{yu2022metaformer} to calculate the mean and variance along both visual token and channel dimensions, as opposed to only channel dimension in vanilla Layer Normalization~\cite{ba2016layer}. MLN can be implemented using GroupNorm API in PyTorch~\cite{paszke2019pytorch} by setting the group number as 1. Our code implementation is based on Pytorch-image-models~\cite{rw2019timm} and MetaFormer baseline~\cite{yu2022metaformer}.

\textbf{Results.} The performance of SPANets on ImageNet classification is presented in Table~\ref{table:acc_imagenet} and Figure~\ref{fig:acc_flops}. Our SPANets outperform others in terms of top-1 accuracy for small, medium, and base models when compared to the CNNs and other MetaFormers based on convolutions or self-attentions. In the case of the small model, SPANet-S achieves better performance than two state-of-the-art MetaFormers, namely LITv2-S and FocalNet-T, despite having a similar number of parameters and FLOPs. Specifically, it outperforms LITv2-S, which uses an attention-based mixer to handle both low and high frequencies, by $1.1\%{p}$, and FocalNet-T, which utilizes a modulated convolution-based mixer, by $0.8\%{p}$. In the medium model case, SPANet-M achieves the highest accuracy with the lowest number of FLOPs and parameters. Even compared to LITv2-M, it gains $0.2\%{p}$ top-1 accuracy. For CNNs, similar to comparison results on small and medium models, SPANets outperform ConvNeXts by $1.0\%{p}$, and $0.4\%{p}$, respectively. Similar results to those of the small and medium cases can also be observed for the base model case.

\begin{table}[]
\caption{\textbf{Performance comparison on ImageNet-1K~\cite{deng2009imagenet} classification.} All models are trained from scratch on the ImageNet-1K training set and the accuracy on the validation set is reported. The numbers of FLOPs for input size $224^{2}$ are counted by \texttt{fvcore}~\cite{fvcore2021} library. The results of RSB-ResNet are from “ResNet Strikes Back”~\cite{wightman2021resnet} which improves the ResNet model~\cite{he2016deep} with an optimized procedure for 300 epochs.}
\centering
\resizebox{\columnwidth}{!}{
\begin{tabular}{l|c|c|c|c|c}
\ChangeRT{1.5pt}
 \multicolumn{1}{c|}{Model}                                   & General Arch.         & Token Mixer        &  Params (M) &  FLOPs (G) & Top-1 (\%) \\ \hline
 RSB-ResNet-50~\cite{he2016deep, wightman2021resnet} & \multirow{2}{*}{CNN} & \multirow{2}{*}{-}  & 26 & 4.1 & 79.8\\
 ConvNeXt-T~\cite{liu2022convnet}                    &                      &                     & 29 & 4.5 & 82.1\\
\cline{2-6}
 PoolFormer-S24~\cite{yu2022metaformer} & \multirow{7}{*}{MetaFormer} &   Pooling & 21 & 3.4 & 80.3 \\
\cline{3-6}
 PVT-Small~\cite{wang2021pyramid} &  & \multirow{3}{*}{Attention}  & 25 & 3.8 & 79.8 \\ 
 Swin-T~\cite{liu2021Swin} &                          &                            &   29 & 4.5 & 81.3 \\
 LITv2-S~\cite{pan2022hilo} &                            &                            &   28 & 3.7 & 82.0 \\
 \cline{3-6}
 GFNet-H-S~\cite{rao2021global} &                            &   \multirow{4}{*}{Convolution}                         &   32 & 4.6 & 81.5 \\
 DWNet-tiny~\cite{han2021connection} &  &  &   24 & 3.8 & 81.2 \\ 
 FocalNet-T~\cite{yang2022focal} &  &  &   29 & 4.5 & 82.3 \\
 \cellcolor{gray!25} SPANet-S (ours) &  &  &   \cellcolor{gray!25}29 & \cellcolor{gray!25}4.6 & \cellcolor{gray!25}\textbf{83.1} \\
\hline
RSB-ResNet-101~\cite{he2016deep, wightman2021resnet} & \multirow{2}{*}{CNN} & \multirow{2}{*}{-} &  45 & 7.9 & 81.3 \\
ConvNeXt-S~\cite{liu2022convnet}         &                      &                    &  50 & 8.7 & 83.1 \\
\cline{2-6}
PoolFormer-M36~\cite{yu2022metaformer} & \multirow{6}{*}{MetaFormer} &   Pooling & 56 & 8.8 & 82.1 \\
\cline{3-6}
PVT-Medium~\cite{wang2021pyramid}                &  & \multirow{3}{*}{Attention} &  44 & 6.7 & 81.2 \\
Swin-S~\cite{liu2021Swin}             &                             &                                 &  50 & 8.7 & 83.0 \\
LITv2-M~\cite{pan2022hilo}               &                             &                                 &  49 & 7.5 & 83.3 \\
\cline{3-6}
 GFNet-H-B~\cite{rao2021global} &                            &   \multirow{3}{*}{Convolution}                         &   54 & 8.6 & 82.9 \\
FocalNet-S~\cite{yang2022focal}          &                             &    &  50 & 8.7 & 83.5 \\
\cellcolor{gray!25} SPANet-M (ours) &  &  &  \cellcolor{gray!25}42 & \cellcolor{gray!25}6.8 & \cellcolor{gray!25}\textbf{83.5} \\
\hline
RSB-ResNet-152~\cite{he2016deep, wightman2021resnet} & \multirow{2}{*}{CNN} &  \multirow{2}{*}{-} &  60 & 11.6 & 81.8 \\
ConvNeXt-B~\cite{liu2022convnet}         &                      &                     &  89 & 15.4 & 83.8 \\
\cline{2-6}
PoolFormer-M48~\cite{yu2022metaformer} & \multirow{8}{*}{MetaFormer} &   Pooling & 73 & 11.6 & 82.5 \\
\cline{3-6}
ViT-B/16~\cite{dosovitskiy2020image} &  & \multirow{4}{*}{Attention} &  86 & 17.6 & 79.7 \\
PVT-Large~\cite{wang2021pyramid} &                          &                            &  61 & 9.8 & 81.7 \\
Swin-B~\cite{liu2021Swin} &                          &                               &  88 & 15.4 & 83.5 \\
LITv2-B~\cite{pan2022hilo} &                            &                            &  87 & 13.2 & 83.6 \\
\cline{3-6}
DWNet-base~\cite{han2021connection} &  & \multirow{3}{*}{Convolution}  &  74 & 12.9 & 83.2 \\
FocalNet-B~\cite{yang2022focal}    &   &                               &  89 & 15.4 & 83.9 \\
\cellcolor{gray!25} SPANet-B (ours) &  &   &  \cellcolor{gray!25}76 & \cellcolor{gray!25}12.0 & \cellcolor{gray!25}\textbf{84.0} \\
\ChangeRT{1.5pt}
\end{tabular}
}
\label{table:acc_imagenet}
\end{table}

\begin{table*}[]
\caption{\textbf{Performance of object detection with RetinaNet~\cite{lin2017focal}, and object detection and instance segmentation with Mask R-CNN~\cite{he2017mask} on COCO val2017~\cite{lin2014microsoft}.} For training detection models, $1\times$ training schedule is adopted consisting of 12 epochs. The performance is reported in terms of bounding box $\text{AP}$ and mask $\text{AP}$, denoted by $\text{AP}^{b}$ and $\text{AP}^{m}$, respectively.}
\centering
\resizebox{\textwidth}{!}{%
\begin{tabular}{l|ccccccc|ccccccc}
\ChangeRT{0.8pt}
\hline
\multicolumn{1}{c|}{}                                 & \multicolumn{7}{c|}{RetinaNet $1\times$}                                                         & \multicolumn{7}{c}{Mask R-CNN $1\times$}                                                             \\ \cline{2-15} 
\multicolumn{1}{c|}{\multirow{-2}{*}{Backbone}}       & \multicolumn{1}{c|}{Param (M)}                  & $\text{AP}$   & $\text{AP}_{50}$ & $\text{AP}_{75}$ & $\text{AP}_{S}$  & $\text{AP}_{M}$  & $\text{AP}_{L}$  & \multicolumn{1}{c|}{Param (M)}                  & $\text{AP}^{b}$  & $\text{AP}^{b}_{50}$ & $\text{AP}^{b}_{75}$ & $\text{AP}^{m}$  & $\text{AP}^{m}_{50}$ & $\text{AP}^{m}_{75}$ \\ \hline
ResNet50~\cite{he2016deep}                                               & \multicolumn{1}{c|}{38}                         & 36.3 & 55.3 & 38.6 & 19.3 & 40.0 & 48.8 & \multicolumn{1}{c|}{44}                         & 38.0 & 58.6  & 41.4  & 34.4 & 55.1  & 36.7  \\
PVT-Small~\cite{wang2021pyramid}                                             & \multicolumn{1}{c|}{34}                         & 40.4 & 61.3 & 43.0 & 25.0 & 42.9 & 55.7 & \multicolumn{1}{c|}{44}                         & 40.4 & 62.9  & 43.8  & 37.8 & 60.1  & 40.3  \\

Swin-T~\cite{liu2021Swin}                                                 & \multicolumn{1}{c|}{39}                         & 41.5 & 62.1 & 44.2 & 25.1 & 44.9 & 55.5 & \multicolumn{1}{c|}{48}                         & 42.2 & 64.6  & 46.2  & 39.1 & 61.6  & 42.0  \\

LITv2-S~\cite{pan2022hilo}                                               & \multicolumn{1}{c|}{38}                         & \textbf{43.7} & -    & -    & -    & -    & -    & \multicolumn{1}{c|}{47}                         & \textbf{44.7} & -     & -     & \textbf{40.7} & -     & -     \\
\rowcolor{gray!25}SPANet-S (ours)                                              & \multicolumn{1}{c|}{\cellcolor{gray!25}38} & 43.3 & \textbf{63.7} & \textbf{46.5} & \textbf{25.8} & \textbf{47.7} & \textbf{57.0} & \multicolumn{1}{c|}{\cellcolor{gray!25}48} & \textbf{44.7} & \textbf{65.7}  & \textbf{48.8}  & 40.6 & \textbf{62.9}  & \textbf{43.8}  \\
\hline
ResNet101~\cite{he2016deep}                                              & \multicolumn{1}{c|}{57}                         & 38.5 & 57.8 & 41.2 & 21.4 & 42.6 & 51.1 & \multicolumn{1}{c|}{63}                         & 40.4 & 61.1  & 44.2  & 36.4 & 57.7  & 38.8  \\
PVT-Medium~\cite{wang2021pyramid}                                             & \multicolumn{1}{c|}{54}                         & 41.9 & 63.1 & 44.3 & 25.0 & 44.9 & 57.6 & \multicolumn{1}{c|}{64}                         & 42.0 & 64.4  & 45.6  & 39.0 & 61.6  & 42.1  \\
Swin-S~\cite{liu2021Swin}                                                 & \multicolumn{1}{c|}{60}                         & 44.5 & \textbf{65.7} & \textbf{47.5} & \textbf{27.4} & \textbf{48.0} & \textbf{59.9} & \multicolumn{1}{c|}{69}                         & 44.8 & \textbf{66.6}  & 48.9  & 40.9 & 63.4  & \textbf{44.2}  \\

LITv2-M~\cite{pan2022hilo}                                               & \multicolumn{1}{c|}{59}                         & \textbf{45.8} & -    & -    & -    & -    & -    & \multicolumn{1}{c|}{68}                         & \textbf{46.5} & -     & -     & \textbf{42.0} & -     & -     \\
\rowcolor{gray!25}
\rowcolor{gray!25}SPANet-M (ours)  & \multicolumn{1}{c|}{\cellcolor{gray!25}51} &   44.0   &   64.3   &   47.0   &   25.9   &  \textbf{48.0}    &   58.7   & \multicolumn{1}{c|}{\cellcolor{gray!25}61} &  45.2    &   66.3    &   \textbf{49.6}    &   41.0   &  \textbf{63.5}     &  44.0     \\  \ChangeRT{0.8pt} \hline
\end{tabular}%
}

\label{table:coco}
\end{table*}

\subsection{Object Detection and Instance Segmentation on COCO}

\textbf{Implementation setup.} SPANet is evaluated based on COCO benchmark~\cite{lin2014microsoft} which includes 118K training images (\texttt{train2017}) and 5K validation images (\texttt{val2017}). The models are trained on the training set, and the performance is reported on the validation set. SPANet is used as the backbone for two widely adopted detectors, namely RetinaNet~\cite{lin2017focal} and Mask R-CNN~\cite{he2017mask}. ImageNet pre-trained weights are used to initialize the backbones, while Xavier initialization~\cite{glorot2010understanding} is utilized to initialize the added layers. All models are trained using AdamW~\cite{kingma2014adam, loshchilov2017decoupled} with an initial learning rate of $1e^{-4}$ and batch size of 8. Following common practices ~\cite{lin2017focal, he2017mask}, we adopted $1\times$ training schedule, which involves training the detection models for 12 epochs. The training images are resized to have a shorter side of 800 pixels, while the longer side is constrained to be at most 1,333 pixels.  For testing, the shorter side of the images is also resized to 800 pixels. The implementation is based on the $\texttt{mmdetection}$~\cite{chen2019mmdetection} codebase.

\textbf{Results.} As shown in Table~\ref{table:coco}, SPANets equipped with RetinaNet~\cite{lin2017focal} show competitive performances compared to their counterparts. For example, SPANet-S achieves 43.3$ \text{AP}$, surpassing ResNet50 (36.3 $\text{AP}$), PVT-Small (40.4 $\text{AP}$), and Swin-T (41.5 $\text{AP}$), while obtaining competitive result to LITv2-S (43.7 $\text{AP}$). Similar results are also observed for SPANet-M. Moreover, these similar results also hold when equipped with Mask R-CNN~\cite{he2017mask}.

\subsection{Semantic Segmentation on ADE20K}

\textbf{Implementation setup.} Following previous studies~\cite{wang2021pyramid, yu2022metaformer}, ADE20K~\cite{zhou2017scene} is selected to benchmark semantic segmentation, which requires an understanding of fine-grained details as well as an ability to analyze long-range interactions. The dataset consists of 20K training and 2K validation images, covering 150 fine-grained categories. We follow the evaluation approach of by employing SPANets as backbones equipped with Semantic FPN~\cite{kirillov2019panoptic} and measuring model performance in terms of mIoU. ImageNet pre-trained weights are adopted to initialize the backbones, while Xavier~\cite{glorot2010understanding} is used to initialize the newly added layers. Following common practices~\cite{kirillov2019panoptic, chen2017deeplab}, models are trained for 80K iterations with a batch size of 16. We employed the AdamW~\cite{kingma2014adam, loshchilov2017decoupled} with an initial learning rate of $2e^{-4}$ that will decay following a polynomial decay schedule with a power of 0.9. Images are randomly resized and cropped into $512\times512$ for training and are rescaled on the shorter side of $512$ pixels for testing. Our code implementation is based on the $\texttt{mmsegmentation}$~\cite{mmseg2020} codebase.

\textbf{Results.} As shown in Table~\ref{table:ADE20K}, equipped with Semantic FPN~\cite{kirillov2019panoptic} for semantic segmentation, SPANet consistently outperforms other existing models. For instance, using nearly identical numbers of parameters and FLOPs, SPANet-S exhibits a $3.9\%{p}$ and $1.1\%{p}$ improvement in mIoU over Swin-T and LITv2-S, respectively. Similar results are also observed for the medium model case.

\begin{table}[]
\caption{\textbf{Performance of semantic segmentation with Semantic FPN~\cite{kirillov2019panoptic} on ADE20K~\cite{zhou2017scene}.} The FLOPs are measured at the resolution of $512\times512$.}
\resizebox{\columnwidth}{!}{%
\begin{tabular}{l|ccc}
\ChangeRT{0.8pt}
\hline
\multicolumn{1}{c|}{{Backbone}} &{Params (M)} & {FLOPs (G)} & {mIoU(\%)} \\ 
\hline
ResNet50~\cite{he2016deep}       & 29                  & 46                 & 36.7              \\ 
PVT-Small~\cite{wang2021pyramid}      & 28                  & 45                 & 39.8              \\
Swin-T~\cite{liu2021Swin}         & 32                  & 46                 & 41.5              \\
LITv2-S~\cite{pan2022hilo}        & 31                  & 41                 & 44.3              \\
\rowcolor{gray!25} SPANet-S (ours)                        & 32                  & 46                 & \textbf{45.4}              \\
\hline
ResNet101~\cite{he2016deep}       & 48                  & 65                 & 38.8              \\ 
PVT-Medium~\cite{wang2021pyramid}      & 48                  & 61                 & 41.6              \\
Swin-S~\cite{liu2021Swin}          & 53                  & 70                 & 45.2              \\
LITv2-M~\cite{pan2022hilo}         & 52                  & 63                 & 45.7              \\
\rowcolor{gray!25} SPANet-M (ours)                       & 45                  & 57                 & \textbf{46.2}             \\
\ChangeRT{0.8pt} \hline
\end{tabular}%
}
\label{table:ADE20K}
\end{table}

\subsection{Ablation}
 
This section presents ablation studies conducted on SPANet using ImageNet-1K~\cite{deng2009imagenet}. The results of these studies are presented in Table~\ref{table:Ablation} and are discussed below according to the following aspects.

\textbf{SPAM components.} To investigate the significance of the components that make up SPAM, we conduct experiments that involve altering the operators. In the first step, it is confirmed the SPF as an important element of SPG. The analysis reveals that the removal of this component results in a significant performance decrease, with accuracy dropping to $82.2\%$. Finally, we find the addition operator is better for context aggregation in SPAM. Our experimental result, shown in Table~\ref{table:Ablation}, indicates that replacing it with the HP leads to a decrease in performance to $82.7\%$.

\textbf{Radius for low-pass band in each stage.} The radius of the low-pass region for each stage is also an important factor affecting performance. As presented in Table~\ref{table:Ablation}, using $[1,1,1,1]$ and $[4,4,1,1]$ decrease the performances in $-0.1\%_{p}$ and $-0.2\%_{p}$, respectively. Therefore, [2,2,1,1] is adopted by default. However, it may not be optimal for SPANet and it is needed to explore optimal parameters to further improve performance in future work.

\textbf{Kernel size for spatial separable convolution.} To examine the kernel size of spatial separable convolution~\cite{Szegedy_2016_CVPR}, we conducted an ablation study using kernels of sizes 3, 5, and 7. Our results indicate that increasing the kernel size from 3 to 7 improves the performance of SPANet from $82.8\%$ to $83.1\%$ while keeping the FLOPs and number of parameters roughly the same. However, we observed that enlarging the kernel from 3 to 5 leads to a decrease in performance. This can be explained by the fact that not all kernels can be split into two separate kernels, which restricts the exploration of all possible kernels and leads to sub-optimal during training. Consequently, we set the kernel size to 7 based on the outcomes of our experiments, \textit{i.e.}, a pair of $1\times7$ and $7\times1$ convolutions is used  by default.

\textbf{Branch output scaling.} The evaluation in the branch output scaling indicates that ResScale~\cite{shleifer2021normformer} is the most effective for SPANet. Notably, when using LayerScale~\cite{touvron2021going}, SPANet exhibits the lowest performance. In other words, we observed that LayerScale~\cite{touvron2021going} has a negative impact on the training of SPANet.

\begin{table}[]
\caption{\textbf{Ablation for SPANet on ImageNet-1K~\cite{deng2009imagenet} classification benchmark.} The number of parameters and FLOPs for all variants are the same, 29 and 4.6 respectively.}
\resizebox{\columnwidth}{!}{%
\begin{tabular}{c|c|c}
\ChangeRT{0.8pt}
\hline
Ablation                                                                                                 & Variant                                                                                        & Top-1(\%)                        \\ \hline
-                                                                                                        & SPANet-S-baseline                                                                              & 82.8                         \\ \hline
\multirow{4}{*}{SPAM components}                                                                         & \multirow{2}{*}{\begin{tabular}[c]{@{}c@{}}SPG with SPF \\ → without SPF\end{tabular}}         & \multirow{2}{*}{82.2 (\textcolor{red}{-0.6})} \\
                                                                                                         &                                                                                                &                              \\ \cline{2-3} 
                                                                                                         & \multirow{2}{*}{\begin{tabular}[c]{@{}c@{}}aggregation with addition\\ → with HP\end{tabular}} & \multirow{2}{*}{82.7 (\textcolor{red}{-0.1})} \\
                                                                                                         &                                                                                                &                              \\ \hline
\multirow{2}{*}{\begin{tabular}[c]{@{}c@{}}Radius for low-\\ pass band in each stage\end{tabular}}  & {[}2,2,1,1{]} → {[}1,1,1,1{]}                                                                  & 82.7 (\textcolor{red}{-0.1})                  \\ \cline{2-3} 
                                                                                                         & {[}2,2,1,1{]} → {[}4,4,1,1{]}                                                                  & 82.6 (\textcolor{red}{-0.2})                  \\ \hline
\multirow{2}{*}{\begin{tabular}[c]{@{}c@{}}Kernel size for spatial\\ separable convolution\end{tabular}} & 3 → 5                                                                                          & 82.7 (\textcolor{red}{-0.1})                  \\ \cline{2-3} 
                                                                                                         & 3 → 7                                                                                          & 83.1 (\textcolor{blue}{+0.3})                  \\ \hline
\multirow{4}{*}{Branch output scaling}                                                                   & \multirow{2}{*}{\begin{tabular}[c]{@{}c@{}}ResScale~\cite{shleifer2021normformer}\\ → None\end{tabular}}            & \multirow{2}{*}{82.7 (\textcolor{red}{-0.1})} \\
                                                                                                         &                                                                                                &                              \\ \cline{2-3} 
                                                                                                         & \multirow{2}{*}{\begin{tabular}[c]{@{}c@{}}ResScale~\cite{shleifer2021normformer}\\ → LayerScale~\cite{touvron2021going}\end{tabular}}      & \multirow{2}{*}{82.6 (\textcolor{red}{-0.2})} \\
                                                                                                         &                                                                                                &                              \\ \ChangeRT{0.8pt} \hline
\end{tabular}%
}
\label{table:Ablation}
\end{table}

%% file: 6_Conclusion.tex
\textbf{Discussion.} In this work, we point out that existing effective token mixers show performance improvements by enhancing either the high- or low-pass filtering capabilities. Based on this, we show that models can be improved using a token mixer that balances of the high- and low-frequency components of the feature map.

To accomplish this, we replace the balancing problem with a mask filtering in the frequency domain and propose SPAM, a novel context aggregation mechanism that enables the optimal balance of high- and low-frequency components for visual features. With SPAM, we build a series of SPANets and evaluate them on three vision tasks. Our experimental results demonstrate that SPANets outperform the state-of-the-art CNNs and MetaFormers based on convolutions or self-attentions for image classification and semantic segmentation. Additionally, SPANets show competitive performances for object detection and instance segmentation.

\textbf{Limitations.} SPANets exhibit limited performance improvements when applied to object detection and instance segmentation tasks. In such dense prediction tasks, identifying the fine-grained details of objects is important and this necessitates utilizing local edges and textures, which correspond to high-frequency components. However, the SPANet backbones, which are pre-trained with ImageNet-1K~\cite{deng2009imagenet}, relatively prioritize low-frequency components to balance frequency components following the Inverse Power Law~\cite{torralba2003statistics}. Consequently, this design choice leads to sub-optimal performance. 

In future work, we will further evaluate SPANets under more different vision tasks which require fine-grained features, such as pose estimation and fine-grained image classification. Moreover, it also requires the development of frequency-balancing token mixers tailored to task-specific characteristics.

%% file: Acknowledgment.tex
This work was supported by the KIST Institutional Program (Project No. 2E32280 and 2E32282), and by the Technology Innovation Program and Industrial Strategic Technology Development Program (20018256, Development of service robot technologies for cleaning a table).

%% file: Supp.tex
\section{Visualizations of Context Aggregation Results from the SPAM}
\begin{figure*}[ht!]
\vspace{0.3cm}
\centering
    \includegraphics[width=0.9\linewidth]{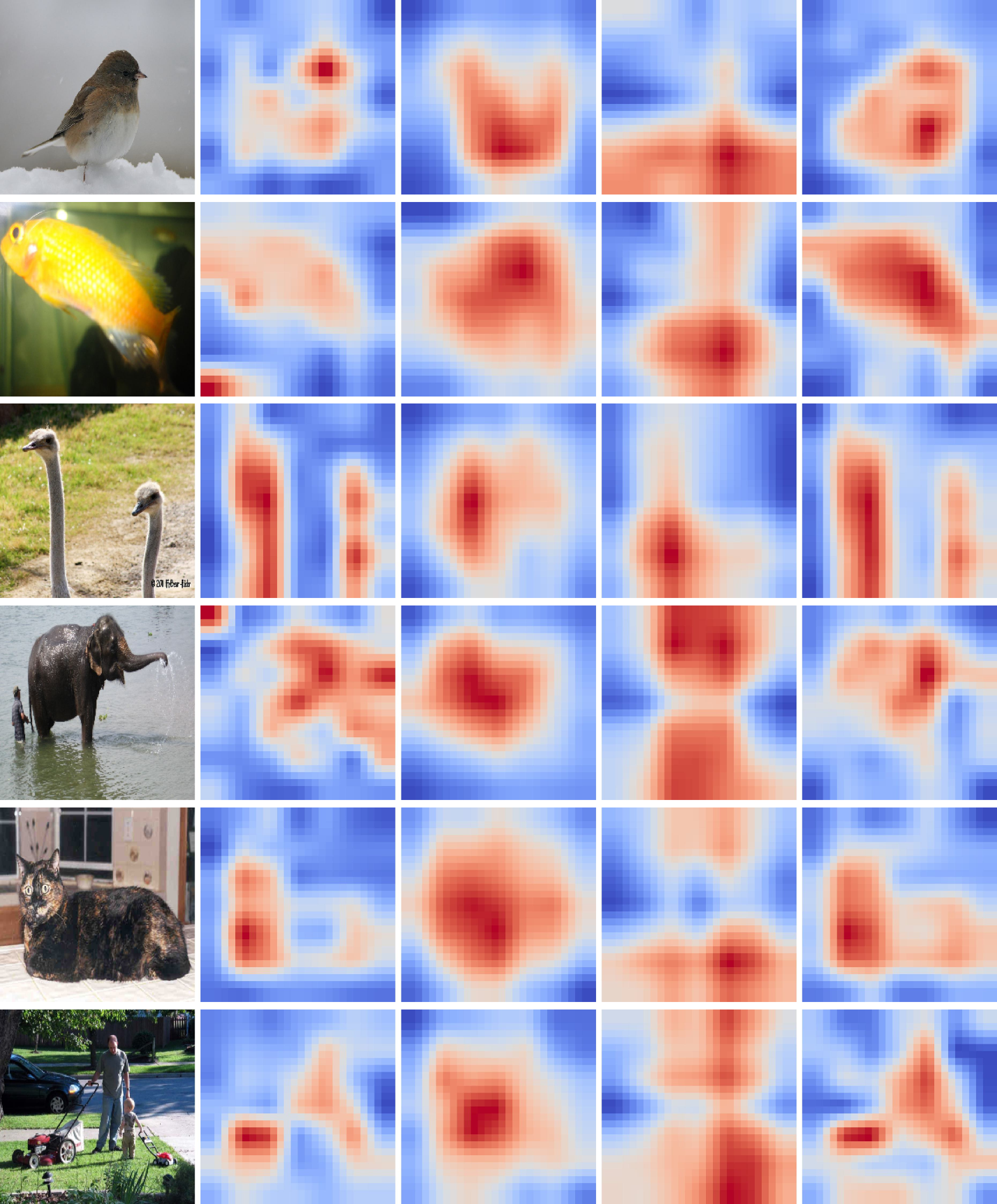}
    \caption{\textbf{Visualization of the results of SPGs and aggregated contexts at the last layer of SPANet-S trained on ImageNet-1K~\cite{deng2009imagenet}.} The columns from left to right are input images, SPG maps at balancing parameters of 0.7, 0.8, and 0.9, and contexts aggregated by addition.}
    \label{fig:vis-spg}
\end{figure*}

Figure~\ref{fig:vis-spg} presents the visualized activation maps of each SPG and the context aggregated by addition, at the last layer of SPANet-S trained on ImageNet-1K~\cite{deng2009imagenet}. The second to fourth columns demonstrate that SPGs learn different contexts at balancing parameters of 0.7, 0.8, and 0.9, respectively. As shown in the second column where the balancing parameter is 0.7 which means relatively weak low-pass filtering (\textit{i.e.}, relatively strong high-pass filtering), SPG concentrates on the overall shapes of objects. The final column depicts the contexts that have been adaptively gathered from the various balancing levels. It shows that the proposed SPAM efficiently captures contextual information of objects, leading to improved performance.

\section{Visualizations of Classification Results}
\begin{figure*}[ht!]
\vspace{0.3cm}
\centering
    \includegraphics[width=0.9\linewidth]{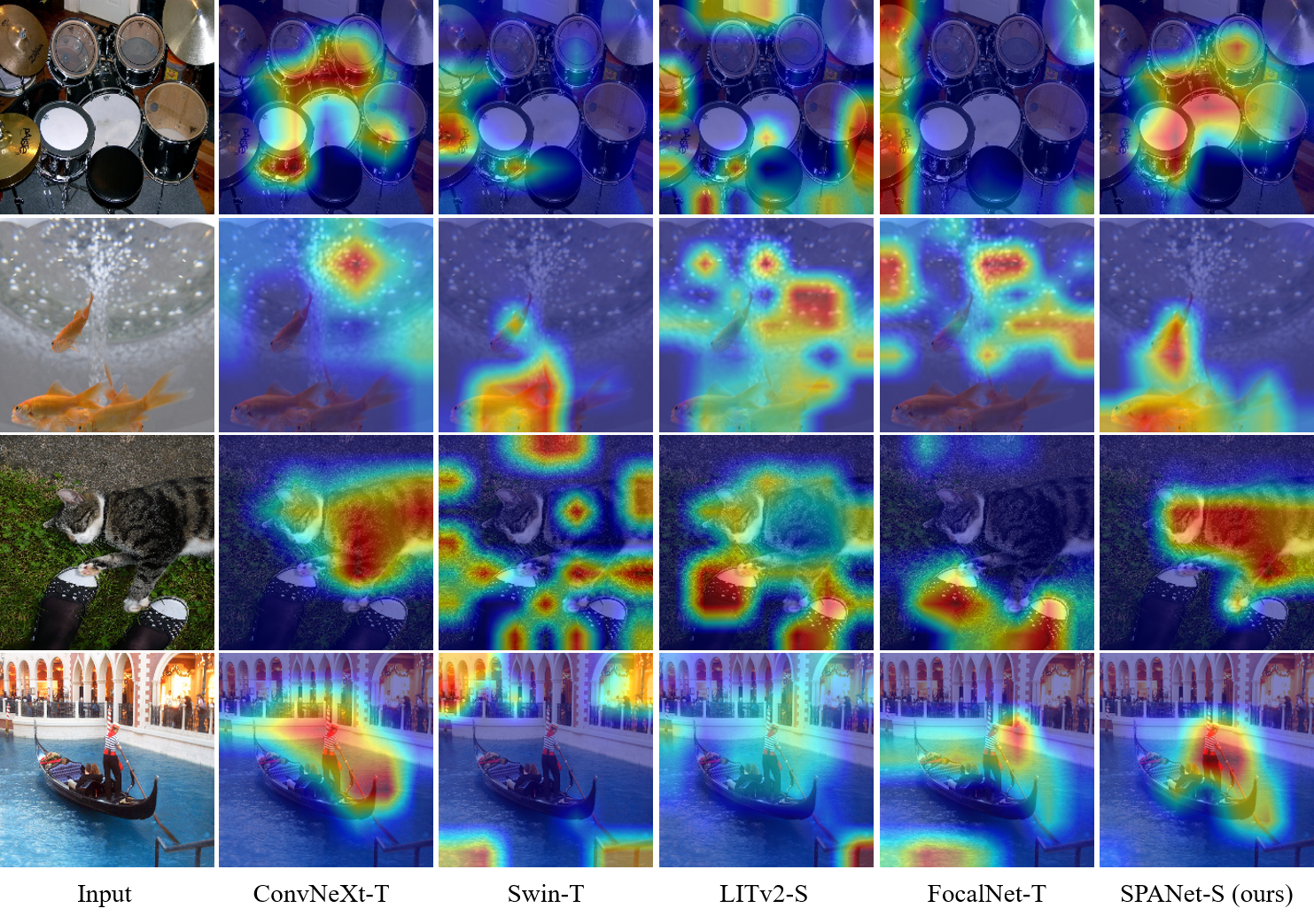}
    \caption{\textbf{Score-CAM~\cite{wang2020score} activation maps of the models trained on ImageNet-1K~\cite{deng2009imagenet}.} The source images are from validation set.}
    \label{fig:score-cam}
\end{figure*}

In order to visualize the results of different models trained on ImageNet-1K~\cite{deng2009imagenet}, we utilized Score-CAM~\cite{wang2020score}. As shown in Figure~\ref{fig:score-cam}, SPANet-S exhibits superior semantic object localization and aggregation capabilities compared to the other models, although all models achieve correct object classification.

\section{Visualizations of Detection and Segmentation Results}
\begin{figure*}[ht!]
\vspace{0.3cm}
\centering
    \includegraphics[width=0.9\linewidth]{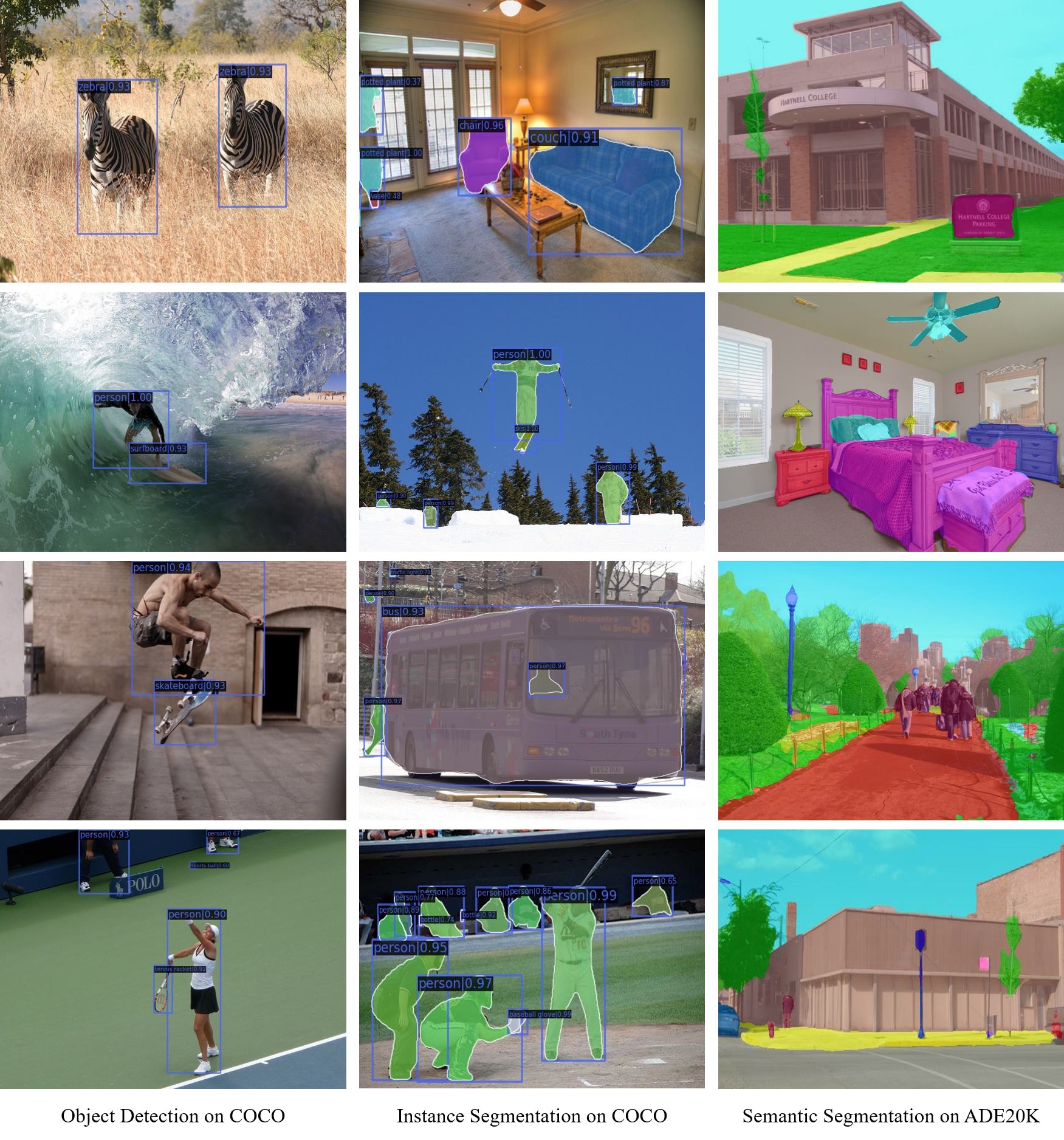}
    \caption{\textbf{Qualitative evaluation results of object detection and instance segmentation on COCO \texttt{val2017}~\cite{lin2014microsoft}, and semantic segmentation on ADE20K~\cite{zhou2017scene}.} The results, ordered from left to right, are generated by SPANet-S backbone equipped with RetinaNet~\cite{lin2017focal}, Mask R-CNN~\cite{he2017mask}, and Semantic FPN~\cite{kirillov2019panoptic}, respectively.}
    \label{fig:coco-dense}
\end{figure*}

In Figure~\ref{fig:coco-dense}, we also present qualitative results for object detection and instance segmentation on COCO \texttt{val2017}~\cite{lin2014microsoft} and semantic segmentation on ADE20K~\cite{zhou2017scene}, showcasing SPANet's ability to integrate seamlessly with dense prediction models like RetinaNet~\cite{lin2017focal}, Mask R-CNN~\cite{he2017mask}, and Semantic FPN~\cite{kirillov2019panoptic}. Our results demonstrate the superior quality of SPANet in achieving high-quality results in these tasks.

